\documentclass[conference]{IEEEtran}
\IEEEoverridecommandlockouts
\usepackage{amsmath,amssymb,amsfonts}
\usepackage{algorithmic}
\usepackage{graphicx}
\usepackage{textcomp}
\usepackage{xcolor}
\usepackage{biblatex}
\usepackage{booktabs}
\usepackage{multirow}
\usepackage{booktabs}
\usepackage[table]{xcolor} 
\usepackage{pifont}
\usepackage{cuted}    
\usepackage{caption}  
\usepackage{array}
\usepackage{arydshln} 
\usepackage[table]{xcolor} 
\def\BibTeX{{\rm B\kern-.05em{\sc i\kern-.025em b}\kern-.08em
    T\kern-.1667em\lower.7ex\hbox{E}\kern-.125emX}}
\begin{document}

\title{MoSoLoco: Monocular 3D Human Localization through Visual Latent Features and Social Context}

\author{
    \IEEEauthorblockN{Yasamin Borhani\textsuperscript{1}, Javad Khoramdel\textsuperscript{2}, Yang Gao\textsuperscript{1}, Po-Chien Luan\textsuperscript{1}, Alexandre Alahi\textsuperscript{1}}
    \IEEEauthorblockA{\textsuperscript{1}Visual Intelligence for Transportation (VITA) Laboratory, EPFL, Switzerland}
    \IEEEauthorblockA{\textsuperscript{2}Faculty of Computer Science, University of Vienna, Vienna, Austria}
    \vspace{-3em}
}
\maketitle

\begin{strip}
    \centering
    \includegraphics[width=0.95\linewidth]{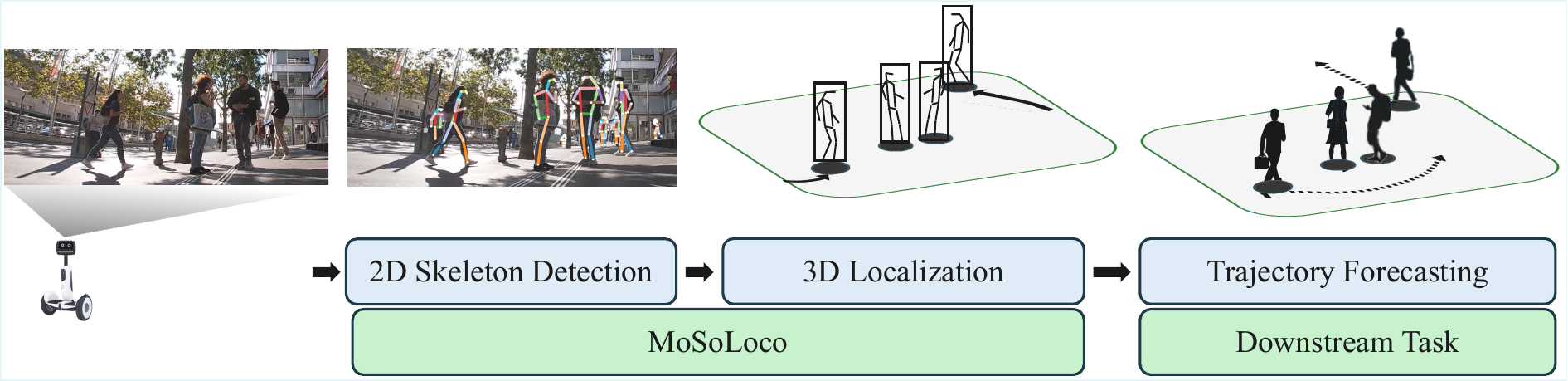}
    \captionof{figure}{Overview of the proposed system architecture. The pipeline processes raw monocular video by extracting 2D human skeletons and joint-aligned latent features from the pose detector. These inputs are fused within our \textbf{MoSoLoco} module, alongside within-frame multi-agent spatial context, to estimate per-frame 3D global coordinates. The coordinates are then evaluated in a downstream trajectory forecasting study.}
    \label{fig:head}
\end{strip}

\begin{abstract}
Estimating the metric 3D position of people from a monocular camera is important for robot navigation and motion prediction. 
Lightweight pose-based localizers are attractive for robotic hardware with limited computational resources because they operate on compact 2D keypoints. However,
they typically discard the richer visual representations computed by the pose estimator.
We introduce \textbf{MoSoLoco}, a per-frame monocular 3D human localizer that combines the pose detector’s internal visual features with the detected 2D skeleton, without requiring an additional RGB or depth encoder. MoSoLoco can further incorporate sparse same-frame multi-person context to refine individual position estimates. 
Experiments on JRDB and NuScenes show that reusing pose-detector representations improves localization, reducing the best JRDB error by approximately 23\% relative
to the strongest evaluated keypoint-based baseline.
A sparse social-interaction module incorporates nearby detections from the same image for further refinement. Controlled trajectory-forecasting experiments show downstream benefits, while the complete pipeline remains suitable for efficient Jetson AGX Xavier inference.

\end{abstract}


\section{Introduction}

\begin{figure*}[t!]
    \centering
    \includegraphics[width=0.95\linewidth]{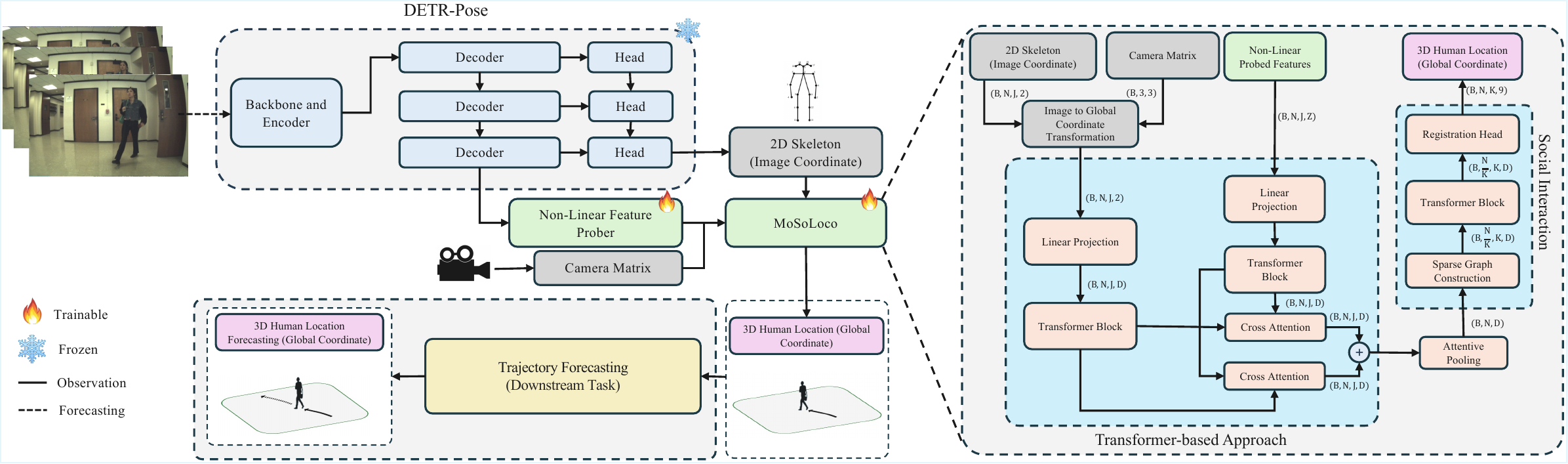}
    \caption{\textbf{Overview of MoSoLoco architecture.} \textbf{(Left)} Given RGB images and camera intrinsic parameters, the framework produces per-frame 3D world coordinates using three key localization components: (1) DETRPose \cite{janampa2025detrpose} for multi-person 2D pose detection, (2) a non-linear probe that maps each joint-specific decoder embedding to a latent feature, and (3) the MoSoLoco network. Fixed trajectory forecasters use the estimated coordinates for downstream evaluation. \textbf{(Right)} MoSoLoco processes explicit 2D skeletons and the joint-aligned latent features via dual streams. Both streams are linearly projected and processed independently before bidirectional cross-attention exchanges geometric and latent information. Attentive pooling produces one token per person hypothesis. An optional sparse $k$-nearest-neighbor attention block then incorporates within-frame context before the regression head recovers the center of the pedestrian's 3D bounding box.}
    \label{fig:overview}
\end{figure*}

Robots operating around people can benefit from real-world 3D human positions for navigation, interaction, and motion prediction  \cite{paez20213d,schmerling2018multimodal,xu2024towards}. Monocular cameras are attractive for compact systems, but perspective projection provides no direct metric depth. One solution is to add a depth sensor such as LiDAR, which increases hardware cost and system complexity; another is to use large monocular depth models such as UniDepth V2~\cite{piccinelli2025unidepthv2}, which provide depth understanding at a considerable computational cost. A pose estimator already provides a compact representation of each person and can support tasks beyond pose estimation, such as pedestrian intention prediction~\cite{razali2021pedestrian}. This motivates us to ask: can computation already performed for human pose estimation also support metric 3D localization?

Pose-based localizers estimate 3D person positions from 2D pose detections. MonoLoco~\cite{bertoni2019monoloco} uses camera-normalized skeletons, while MonoTransmotion~\cite{luan2025unified} adds temporal pose observations. In these approaches, localization relies mainly on the final keypoints, even though the detector computes richer visual representations before producing them. Similarly, trajectory-forecasting methods model interactions among nearby people~\cite{alahi2016social,saadatnejad2024socialtransmotion}, whereas pose-based localizers have generally focused on the target person and temporal observations rather than same-frame spatial context. These observations point to two underused sources of localization information: intermediate detector features
and neighboring detections.

The study proceeds in two stages. First, a controlled feature probe tests whether frozen pose-detector representations retain camera-depth information beyond coarse skeleton-derived cues. These observations motivate \textbf{MoSoLoco}, a per-frame monocular 3D human localizer that estimates each person’s 3D position from the detector’s visual
representations and the detected 2D skeleton expressed in camera coordinates, as illustrated in Fig.~\ref{fig:head}. The pose detector remains frozen, so no additional RGB or depth encoder is required. Two localizer designs are considered: a lightweight MLP-based model and a more expressive Transformer-based model. For multi-person scenes, the model can optionally use sparse attention over nearby detections from the same image. This module uses neither identities nor observations from previous frames, allowing the contributions of detector representations and same-frame spatial context
to be evaluated independently.
  
We evaluate the method on JRDB~\cite{martin2021jrdb} and NuScenes~\cite{caesar2020nuscenes}. The experiments separate the effects of skeleton geometry, pose-detector representations, fusion architecture, detector capacity, and same-frame multi-person context. We also evaluate the estimated coordinates as observations for fixed trajectory forecasters, following the view that errors in real perception inputs can propagate to downstream motion prediction~\cite{xu2024towards}. Finally, we measure detector--localizer inference on an embedded robotic platform.

Our contributions are as follows:
\begin{itemize}
\item \textbf{Visual cues for monocular 3D localization:} a controlled analysis shows that visual features computed inside a frozen 2D pose detector retain distance-related information beyond coarse skeleton geometry. MoSoLoco incorporates these features with the detected 2D skeleton to estimate per-person 3D positions, without adding a separate image or depth encoder.

\item \textbf{Multi-person spatial context:} inspired by the value of inter-person context in human motion modeling, we investigate whether neighboring detections can also refine per-person monocular localization and introduce a sparse attention module that remains frame-based.

\item \textbf{Empirical validation across settings:} across JRDB and NuScenes with two detector capacities, MoSoLoco reduces localization error by up to 23.9\% relative to the evaluated keypoint-based baselines. On JRDB, it matches or exceeds the accuracy of the much larger UniDepth and FCOS3D reference systems, while downstream forecasting and embedded-deployment studies assess its practical utility.
\end{itemize}

\section{Related Work}

Monocular 3D localization is commonly addressed through 3D object detection~\cite{wang2021fcos3d,wang2021depth,huang2022monodtr,wu2023monopgc} or metric-depth estimation~\cite{piccinelli2025unidepthv2}. These approaches provide general image-to-3D solutions but require a dedicated visual or depth network. The setting considered here is different: a human-centric perception stack already contains a 2D pose estimator, and the question is whether representations computed inside that estimator can also provide metric localization cues without introducing another image encoder.

Pose-based methods offer a lightweight alternative by exploiting human geometry. MonoLoco~\cite{bertoni2019monoloco} maps camera-normalized 2D skeletons to 3D locations, while MonStereo~\cite{bertoni2021monstereo} and pose-aware localization methods~\cite{rahimi2021poseaware} further exploit human priors and visibility cues. Chen et al.~\cite{chen2023pedestrianfusion} complement skeleton-based localization with pseudo-LiDAR depth from an auxiliary image network, whereas MonoTransmotion~\cite{luan2025unified} uses temporal pose sequences to improve localization and trajectory prediction. MoSoLoco instead reuses joint-aligned features already computed by the pose detector and fuses them with skeleton geometry. It requires neither an additional depth/image network nor temporal observations, and it additionally investigates sparse same-frame multi-person context.

Human trajectory forecasting has evolved from recurrent social models~\cite{alahi2016social} to generative and attention-based approaches~\cite{gupta2018social,ivanovic2020multimodal,saadatnejad2024socialtransmotion,gao2025omnitraj}. The contribution here is not a new forecasting model. Forecasting serves as a controlled downstream evaluation of localization quality: fixed predictors and ground-truth association isolate whether improved per-frame 3D observations produce more useful observed trajectories.

\section{Methodology}

MoSoLoco estimates the metric 3D center of each detected person from a single frame while reusing representations already computed by the upstream pose estimator. As shown in Fig.~\ref{fig:overview}, each DETRPose~\cite{janampa2025detrpose} detection provides two complementary representations: explicit 2D skeleton geometry and joint-aligned decoder features available before keypoint regression. MoSoLoco fuses these representations into a person-level embedding and optionally refines it using sparse context from other people in the same frame. DETRPose remains frozen during localizer training, so no additional RGB or depth encoder is introduced.

\subsection{Skeleton Geometry and Pose-Feature Reuse}
For person $i$ and joint $j$, let $(u_{ij},v_{ij})$ denote the detected image coordinates. Camera-dependent pixel scale is removed using intrinsics $(f_x,f_y,c_x,c_y)$:
\begin{equation}
\mathbf{g}_{ij}=\left[\frac{u_{ij}-c_x}{f_x},\frac{v_{ij}-c_y}{f_y}\right]^T.
\end{equation}
The resulting $J=17$ joint tokens retain the explicit geometry used by conventional pose-based localizers.

Rather than discarding the detector representation after keypoint prediction, the corresponding 256-dimensional DETRPose embedding $\mathbf{h}_{ij}$ is reused for each joint. A three-layer non-linear probe $\phi(\cdot)$ maps it to $\mathbf{l}_{ij}=\phi(\mathbf{h}_{ij})\in\mathbb{R}^{128}$. The probe is trained jointly with the localizer while DETRPose remains frozen, allowing the visual representation to be reused without a second image backbone.

In \textbf{MoSoLoco-Trans}, geometry and latent features are projected to 256-dimensional joint tokens and processed independently before bidirectional cross-attention exchanges information between the two streams. Residual fusion followed by attentive pooling produces a 128-dimensional person token. This preserves the joint-level correspondence between explicit geometry and detector features while allowing either representation to complement ambiguities in the other.

\textbf{MoSoLoco-MLP} tests whether the benefit of feature reuse depends on Transformer fusion. The 34-dimensional flattened skeleton is concatenated with the mean-pooled 128-dimensional latent representation, projected to a 256-dimensional embedding, and processed by three residual MLP stages. Both variants therefore use the same geometric and detector information but differ in how those representations are fused.

\subsection{Sparse Same-Frame Multi-Person Context}
\label{subsec:social_interaction}
Monocular depth can remain ambiguous for an isolated person hypothesis. Each person representation can therefore be refined using other detections visible in the same frame. For each hypothesis, the center of its 2D skeleton is computed in the normalized camera plane and the $k_i=\min(K,Q-1)$ nearest external hypotheses are retained, where $Q$ is the number of detected people. The person's own token is included in the neighborhood.

A single multi-head attention and feed-forward block updates the representation from these sparse neighbors. Restricting attention to at most $K$ external hypotheses reduces person-level attention from $O(Q^2D)$ to $O(Q(K+1)D)$ for token dimension $D$. The module is strictly per-frame: it uses neither identities nor observations from previous frames and therefore does not turn MoSoLoco into a temporal localization or tracking method.

\subsection{3D Localization and Training Objective}
The prediction target is the center $\mathbf{c}=(x_c,y_c,z_c)$ of the annotated 3D pedestrian bounding box in the camera frame. Following MonoLoco~\cite{bertoni2019monoloco}, its position is parameterized by azimuth $\theta$, polar angle $\psi$, and radial distance $r=\lVert\mathbf{c}\rVert_2$. The regression head predicts $[\theta,\psi,r,s,h,w,l,\sin\alpha,\cos\alpha]$, where $s$ parameterizes Laplacian uncertainty in distance, $(h,w,l)$ are normalized box dimensions, and $\alpha$ is the observation angle. Dimensions and orientation provide auxiliary supervision; only $(\theta,\psi,r)$ determine the evaluated center.

For matched prediction--target pairs $\mathcal{M}$, the localizer is trained with
\begin{equation}
\begin{aligned}
\mathcal{L}_{\mathrm{loc}}=\Big\langle&
\left|1-\frac{\hat r}{r}\right|e^{-\hat s}+\hat s
+\!\sum_{q\in\{\theta,\psi,h,w,l\}}\!|\hat q-q|\\
&+\tfrac12\left(|\hat a-\sin\alpha|+|\hat b-\cos\alpha|\right)
\Big\rangle_{\mathcal{M}},
\end{aligned}
\end{equation}
where $\hat a$ and $\hat b$ are the predicted sine and cosine components of the observation angle. All task terms have unit weight. The pose estimator remains frozen, and no additional pose-detection or localization-ADE loss is used.

At inference, the camera-frame center is reconstructed as
\begin{equation}
\hat{\mathbf{c}}=
\begin{bmatrix}
\hat r\sin\hat\psi\cos\hat\theta &
\hat r\cos\hat\psi &
\hat r\sin\hat\psi\sin\hat\theta
\end{bmatrix}^{T},
\end{equation}
and transformed to the global frame using the known camera and robot extrinsics. These global ground-plane centers are used for localization evaluation and, without changing the localizer, as observations in the controlled forecasting study of Section~\ref{subsec:trajectory_forecasting}.

\section{Experiments}

The experiments are organized around the central claim that a pose detector contains useful localization cues beyond its final joint coordinates. JRDB provides the main localization comparison and component analyses, NuScenes evaluates the approach in a second camera and scene domain, and controlled forecasting and embedded-deployment studies examine downstream value and practical use.

\subsection{Localization Protocols}
\textbf{JRDB:} The evaluation split contains 1,658 images and 3,422 annotated person-frame targets before filtering. MoSoLoco, MonoLoco~\cite{bertoni2019monoloco}, and MonoTransmotion~\cite{luan2025unified} are trained for 100 epochs with batch size 16 and learning rate $5\times10^{-4}$. Detector hypotheses are filtered at confidence 0.20 and pose-box NMS IoU 0.50, then matched to annotations using Hungarian assignment with minimum OKS 0.50. ADE is the mean Euclidean error between predicted and annotated pedestrian centers in the global ground plane over accepted matches. Coverage is the fraction of annotated targets retained by this policy. DETRPose-S retains 3,162/3,422 targets (92.40\%), while DETRPose-N retains 2,794/3,422 (81.65\%).

\textbf{NuScenes:} NuScenes does not provide human keypoints, so preprocessing follows MonoTransmotion~\cite{luan2025unified}. OpenPifPaf~\cite{kreiss2019pifpaf} produces pseudo poses in COCO-17 order that are associated with pedestrian annotations. DETRPose-S is initialized from its official COCO weights and its Transformer pose head is adapted for 10 epochs on pseudo poses from the official training scenes. Each localizer is then trained for 100 epochs on the same training scenes with fresh optimizer and scheduler states. Evaluation uses the official validation scenes, comprising 2,496 valid person-frame targets with mean and median camera-to-person distances of 16.13 m and 14.95 m. Confidence, NMS, and OKS thresholds match JRDB.

\textbf{External systems:} UniDepth V2 ViT-L~\cite{piccinelli2025unidepthv2} and FCOS3D R101~\cite{wang2021fcos3d} provide broader reference points rather than controlled localizer ablations. On JRDB, frozen UniDepth uses DETRPose-S hypotheses and the same matching policy. Its calibrated variant applies a fixed per-camera ray correction estimated on the training split and a $+0.14$ m validation-selected depth offset; evaluation annotations are not used for calibration. FCOS3D is initialized from its official NuScenes-pretrained checkpoint and selected after one epoch of JRDB fine-tuning. On NuScenes, UniDepth again uses DETRPose-S hypotheses, whereas FCOS3D performs its own detection. Because detection, training, and depth-processing pipelines differ, these systems are reported as contextual accuracy--resource comparisons rather than parameter-matched ablations.

\textbf{Resource measurements:} FP32 inference costs are measured on an NVIDIA A100-SXM4-80GB at batch size one. Peak memory is the maximum \texttt{torch.cuda.max\_memory\_allocated} over three forward passes after two warm-up passes and includes parameters, inputs, activations, outputs, and temporary PyTorch allocations. GFLOPs are measured from the executed graph with CalFlops. Learned localizers use $Q=5$ people per image, 17 joints, 128-dimensional detector latents, and $K=4$ unless stated otherwise. DETRPose uses $640\times640$ input, and the shared detector cost is listed separately from each localizer.

\subsection{Controlled Forecasting Protocol}
\label{subsec:trajectory_forecasting}
Trajectories are sampled at 2.5 Hz. Each example contains eight observation slots (3.2 s), of which two to eight may contain accepted localizer predictions, followed by 12 prediction frames (4.8 s). The Vanilla Transformer and OmniTraj~\cite{gao2025omnitraj} are each trained once using only ground-truth locations and identities, with batch size 16 and learning rate $10^{-4}$. At evaluation time, every localizer is paired with the same trained forecaster and the same ground-truth future trajectories.

Ground-truth identities are used only as an offline association control. In each observed frame, post-filtering pose detections are assigned one-to-one to ground-truth people by Hungarian keypoint matching, and only matches satisfying the OKS threshold are retained. Accepted detections inherit the corresponding identity for this experiment; rejected detections remain missing and are represented with a zero temporal mask. A window is scored when the final observed frame is valid and at least two of the eight observation slots are valid. Methods are compared on the intersection of retained windows within each detector condition.

Forecasting ADE is the mean ground-plane error over the 12 future steps and FDE is the error at the final future step. OmniTraj uses best-of-20 minimum ADE and the FDE from the same ADE-selected mode. The DETRPose-S common cohort retains 1,277 of 1,349 forecast-scored ground-truth-eligible trajectories (94.66\%) across 631 cache scenes. DETRPose-N retains 1,054 trajectories across 568 cache scenes (78.13\%). Coverage is reported with ADE/FDE because missed or rejected observations are not assigned zero displacement error.

For the filtered-history condition, only the observed prefix is processed with a second-order forward--backward Butterworth filter ($W_n=1/6$). Missing slots are interpolated only for filtering and restored to missing afterward; masks and future coordinates remain unchanged. The filter therefore uses no information after the forecast origin.

\subsection{Embedded Deployment Protocol}
DETRPose-S with the Transformer and MLP social localizers ($K=4$) is evaluated on an NVIDIA Jetson AGX Xavier using JetPack 5.1.2, TensorRT 8.5.2, batch size one, and $640\times640$ inputs. The board runs in 30 W all-core mode with fixed CPU, GPU, and memory clocks. The baseline uses PyTorch FP32. The optimized implementation uses a mixed FP16/FP32 TensorRT detector, FP32 TensorRT localizers, reused TensorRT buffers, and an exact stable-greedy CUDA NMS kernel. Both runtimes use identical checkpoint weights and the same 1,658-image JRDB evaluation payload.

Latency is measured with CUDA synchronization around each forward pass; the first 10 images are excluded as warm-up. The timed region contains DETRPose inference, confidence filtering, NMS, and localization. Image loading and resizing, host-to-device transfer, and final metric post-processing are excluded. Reported FPS therefore measures detector--localizer model inference rather than full sensor-to-output system throughput.

\section{Results}

\begin{table}[t]
\centering
\caption{Frozen DETRPose feature association with camera-axis depth. PC is the strongest absolute Pearson-correlated component among the first 32; EV is explained variance. Partial $r$ controls the coarse cues described in text.}
\label{tab:detector_depth_correlation}
\small
\setlength{\tabcolsep}{2pt}
\begin{tabular}{@{}lccccc@{}}
\toprule
\textbf{Dataset} & \textbf{Matched} & \textbf{PC (EV)} & \textbf{Pearson $r$} & \textbf{Spearman $\rho$} & \textbf{Partial $r$} \\ \midrule
JRDB & 2,964 & 3 (11.2\%) & 0.652 & 0.643 & 0.168 \\
NuScenes & 2,252 & 1 (15.0\%) & 0.846 & 0.852 & 0.369 \\ \bottomrule
\end{tabular}
\end{table}

\begin{table*}[t]
\centering
\caption{JRDB localization. Params/Size report localizer parameters and FP32 storage; Peak/GFLOPs follow the resource protocol. Detector headers give shared front-end cost. Bold denotes best ADE.}
\label{tab:localization_results}
\scriptsize
\setlength{\tabcolsep}{3pt}
\begin{tabular}{@{}lcccccccc@{}}
\toprule
\textbf{Localization Model} & \textbf{Skeleton} & \textbf{Latent} & \textbf{Social} & \textbf{Params (M)} & \textbf{Size (MB)} & \textbf{Peak (MB)} & \textbf{GFLOPs} & \textbf{ADE $\downarrow$} \\ \midrule
\multicolumn{9}{@{}l}{\cellcolor{black!5} \textbf{End-to-end monocular 3D detector}} \\ \midrule
FCOS3D R101 \cite{wang2021fcos3d} & \ding{55} & \ding{55} & \ding{55} & 55.0 & 219.8 & 430.9 & 169.46 & 0.550 \\ \midrule
\multicolumn{9}{@{}l}{\cellcolor{black!5} \textbf{Detector:} DETRPose-S \cite{janampa2025detrpose} (11.5 M, 46.9 MB; peak: 146.7 MB; 32.42 GFLOPs; COCO AP: 67.0; coverage: 92.40\%)} \\ \midrule
MonoLoco \cite{bertoni2019monoloco} & \ding{51} & \ding{55} & \ding{55} & 0.41 & 1.6 & 11.3 & 0.004 & 0.326 \\
MonoTransmotion \cite{luan2025unified} & \ding{51} & \ding{55} & \ding{55} & 5.29 & 21.2 & 30.8 & 0.053 & 0.322 \\
UniDepth V2 ViT-L (uncalibrated) \cite{piccinelli2025unidepthv2} & \ding{51} & \ding{55} & \ding{55} & 353.8 & 1415.3 & 2173.2 & 1612.09 & 0.765 \\
UniDepth V2 ViT-L (calibrated) \cite{piccinelli2025unidepthv2} & \ding{51} & \ding{55} & \ding{55} & 353.8 & 1415.3 & 2173.2 & 1612.09 & 0.622 \\ \hdashline \noalign{\vskip 0.5ex}
\textbf{MoSoLoco-MLP (latent-only)} & \ding{55} & \ding{51} & \ding{55} & 0.43 & 1.7 & 11.4 & 0.004 & 0.289 \\
\textbf{MoSoLoco-MLP} & \ding{51} & \ding{51} & \ding{55} & 0.44 & 1.8 & 11.4 & 0.004 & 0.266 \\
\textbf{MoSoLoco-MLP} & \ding{51} & \ding{51} & \ding{51} & 0.67 & 2.7 & 12.4 & 0.008 & 0.254 \\
\textbf{MoSoLoco-Trans (latent-only)} & \ding{55} & \ding{51} & \ding{55} & 2.05 & 8.2 & 18.6 & 0.107 & 0.267 \\
\textbf{MoSoLoco-Trans} & \ding{51} & \ding{51} & \ding{55} & 4.42 & 17.7 & 29.0 & 0.465 & 0.262 \\
\textbf{MoSoLoco-Trans} & \ding{51} & \ding{51} & \ding{51} & 4.62 & 18.5 & 29.8 & 0.468 & \textbf{0.248} \\ \midrule
\multicolumn{9}{@{}l}{\cellcolor{black!5} \textbf{Detector:} DETRPose-N \cite{janampa2025detrpose} (4.1 M, 16.8 MB; peak: 73.3 MB; 9.14 GFLOPs; COCO AP: 57.2; coverage: 81.65\%)} \\ \midrule
MonoLoco \cite{bertoni2019monoloco} & \ding{51} & \ding{55} & \ding{55} & 0.41 & 1.6 & 11.3 & 0.004 & 0.333 \\
MonoTransmotion \cite{luan2025unified} & \ding{51} & \ding{55} & \ding{55} & 5.29 & 21.2 & 30.8 & 0.053 & 0.343 \\ \hdashline \noalign{\vskip 0.5ex}
\textbf{MoSoLoco-MLP (latent-only)} & \ding{55} & \ding{51} & \ding{55} & 0.43 & 1.7 & 11.4 & 0.004 & 0.351 \\
\textbf{MoSoLoco-MLP} & \ding{51} & \ding{51} & \ding{55} & 0.44 & 1.8 & 11.4 & 0.004 & 0.297 \\
\textbf{MoSoLoco-MLP} & \ding{51} & \ding{51} & \ding{51} & 0.67 & 2.7 & 12.4 & 0.008 & 0.276 \\
\textbf{MoSoLoco-Trans (latent-only)} & \ding{55} & \ding{51} & \ding{55} & 2.05 & 8.2 & 18.6 & 0.107 & 0.347 \\
\textbf{MoSoLoco-Trans} & \ding{51} & \ding{51} & \ding{55} & 4.42 & 17.7 & 29.0 & 0.465 & 0.273 \\
\textbf{MoSoLoco-Trans} & \ding{51} & \ding{51} & \ding{51} & 4.62 & 18.5 & 29.8 & 0.468 & \textbf{0.265} \\ \bottomrule
\end{tabular}
\end{table*}

The results progress from analyzing the information used by MoSoLoco to evaluating the complete detector--localizer pipeline. The first analysis characterizes the
depth-related structure retained in frozen pose-detector features. The JRDB study then quantifies the resulting localization gains, with component ablations
separating the contributions of detector representations and multi-person context. NuScenes extends the evaluation to a different camera and scene domain, while
the final experiments examine downstream trajectory forecasting and embedded inference efficiency.

\subsection{Depth-Related Structure in Frozen Pose-Detector Features}
\label{subsec:detector_depth_correlation}
A descriptive probe tests whether frozen DETRPose features retain distance-related information before localization. The 17 decoder embeddings of each matched detection are mean-pooled to 256 dimensions; PCA is applied and correlations with camera-axis depth are evaluated over the first 32 components. Table~\ref{tab:detector_depth_correlation} reports the component with the strongest absolute Pearson correlation. Partial $r$ additionally regresses both depth and that component on coarse skeleton/camera cues (size, spread, position, aspect ratio, confidence, normalized focal length, and scale-derived terms).

The raw correlations are substantial on JRDB ($r=0.652$, $\rho=0.643$) and NuScenes ($r=0.846$, $\rho=0.852$), but partial $r$ falls to 0.168 and 0.369. Thus, geometry explains much of the association, while a positive residual association remains beyond the controlled coarse cues. No localizer outputs enter this probe; subsequent ablations test task utility.

\subsection{JRDB Localization}

Table~\ref{tab:localization_results} gives the main JRDB comparison. With DETRPose-S, social MoSoLoco-Trans reaches 0.248 m ADE, a 23.9\% reduction from MonoLoco (0.326 m) and 23.0\% from MonoTransmotion (0.322 m); with DETRPose-N, it reaches 0.265 m versus 0.333/0.343 m. The social MLP obtains 0.254/0.276 m for S/N. FCOS3D and UniDepth use different pipelines and are contextual references: FCOS3D obtains 0.550 m over 3,271/3,422 targets (95.59\%), while UniDepth obtains 0.765 m uncalibrated and 0.622 m calibrated on 2,964/3,422 targets (86.62\%).

\subsection{Component Analysis}
\textbf{Pose-detector features:} Without social context, geometry+latent MLP reduces ADE from MonoLoco's 0.326 to 0.266 m with DETRPose-S and from 0.333 to 0.297 m with DETRPose-N. Latent-only MLP/Transformer models reach 0.289/0.267 m (S) and 0.351/0.347 m (N); restoring skeleton geometry gives 0.266/0.262 m and 0.297/0.273 m, respectively.

\textbf{Same-frame context:} Social attention improves all four detector--architecture point estimates: MLP changes from 0.266 to 0.254 m (S) and 0.297 to 0.276 m (N), while Transformer changes from 0.262 to 0.248 m and 0.273 to 0.265 m.

\begin{figure*}[t]
    \centering
    \begin{minipage}[t]{0.49\textwidth}
        \centering
        \includegraphics[width=\linewidth]{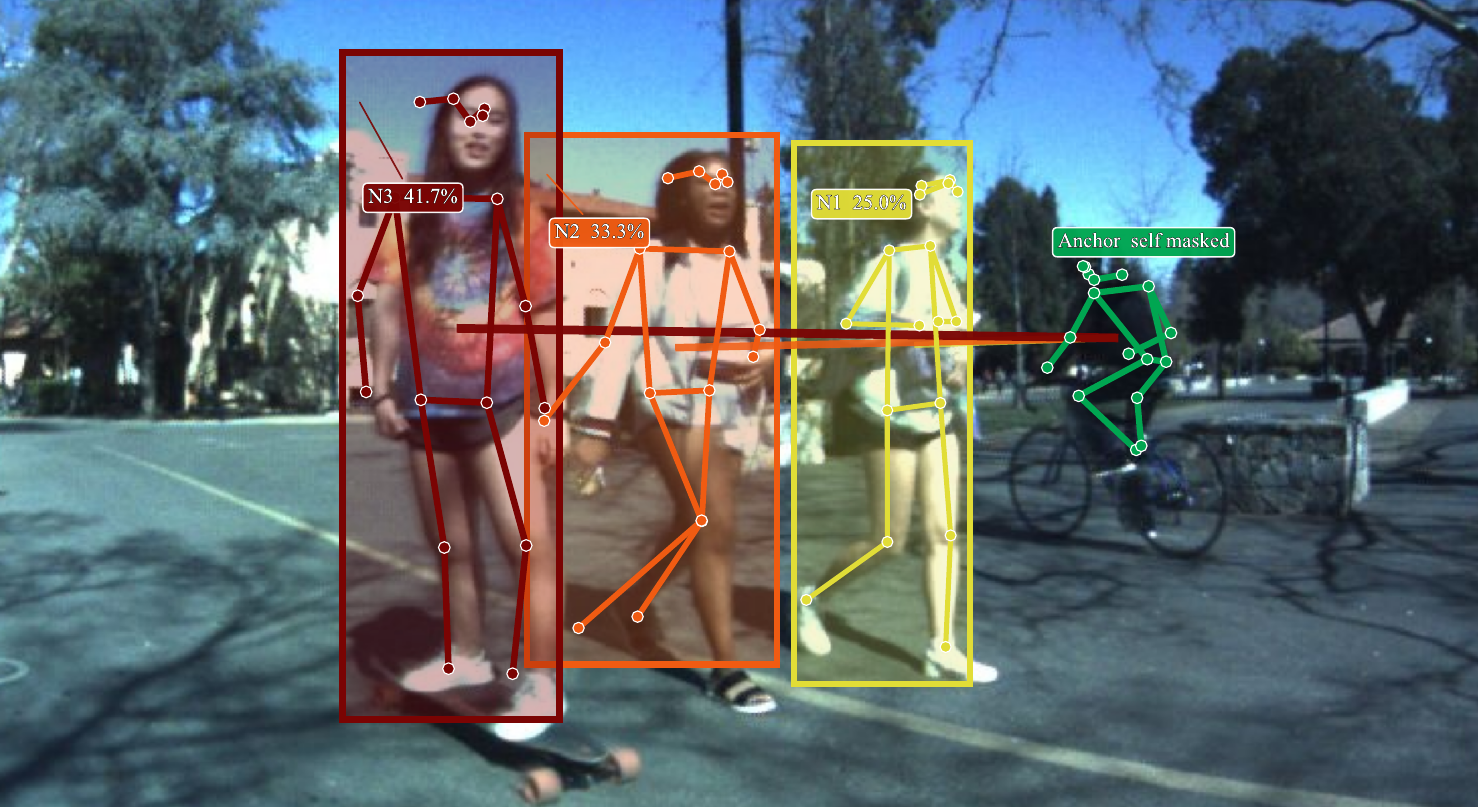}
        \vspace{-1mm}
        {\small (a) Social refinement degrades the estimate.}
    \end{minipage}
    \hfill
    \begin{minipage}[t]{0.48\textwidth}
        \centering
        \includegraphics[width=\linewidth]{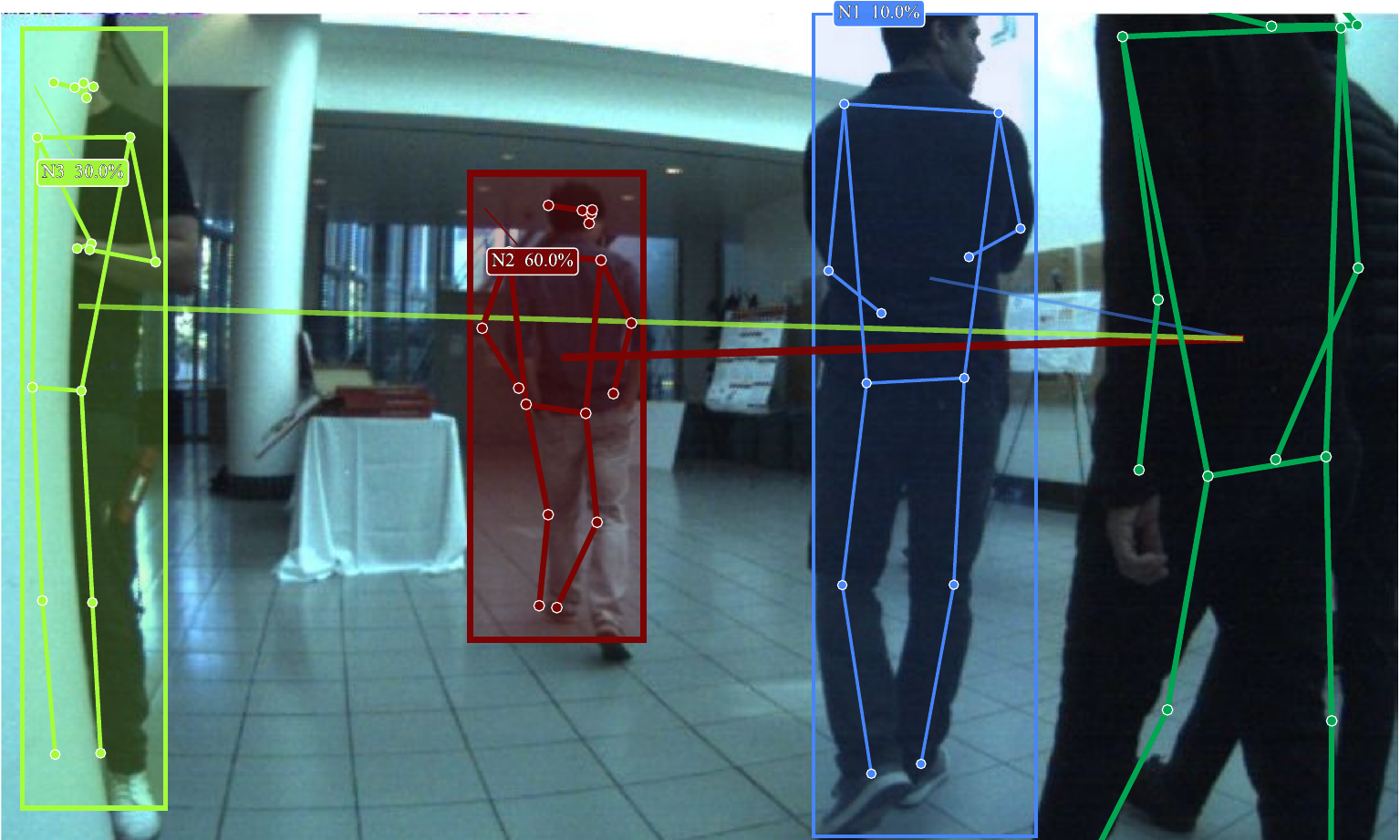}
        \vspace{-1mm}
        {\small (b) Social refinement improves the estimate.}
    \end{minipage}
    \caption{\textbf{Qualitative social refinement.} The anchor is shown with attended neighbors; color denotes relative attention weight. \textbf{(a)} A compact, strongly articulated anchor overlaps a bicycle while neighbors are upright pedestrians, and the social variant is less accurate. \textbf{(b)} For an anchor truncated by the image boundary, social context improves localization. Attention weights indicate allocation, not causal importance.}
    \label{fig:social_examples}
\end{figure*}

\begin{table}[t]
\centering
\caption{Maximum external-neighbor ablation for MoSoLoco-Trans.}
\label{tab:k_ablation}
\small
\setlength{\tabcolsep}{3pt}
\begin{tabular}{@{}>{\raggedright\arraybackslash}m{0.38\columnwidth} >{\centering\arraybackslash}m{0.28\columnwidth} >{\centering\arraybackslash}m{0.28\columnwidth}@{}}
\toprule
\multicolumn{3}{@{}>{\raggedright\arraybackslash}p{0.94\columnwidth}@{}}{\cellcolor{black!5} \textbf{Detector:} DETRPose-S} \\
\multicolumn{3}{@{}>{\raggedright\arraybackslash}p{0.94\columnwidth}@{}}{\cellcolor{black!5} JRDB coverage: 92.40\%; NuScenes coverage: $\sim$90.1\%} \\ \midrule
\textbf{Social setting} & \textbf{JRDB ADE $\downarrow$} & \textbf{NuScenes ADE $\downarrow$} \\ \midrule
Without social & 0.262 & 1.427 \\ \hdashline \noalign{\vskip 0.5ex}
$K=1$ & 0.252 & 1.146 \\
$K=2$ & 0.261 & 1.185 \\
$K=4$ & \textbf{0.248} & \textbf{1.130} \\
$K=6$ & 0.252 & 1.181 \\
\bottomrule
\end{tabular}
\end{table}

Table~\ref{tab:k_ablation} shows a non-monotonic effect of neighborhood size; every tested $K$ improves on no social context, and $K=4$ is best on both datasets. Because $K$ is an upper bound, each frame uses $\min(K,Q-1)$ external neighbors. Figure~\ref{fig:social_examples} illustrates complementary cases: context is less reliable for a strongly articulated anchor that differs from upright neighbors, but helps when the anchor is heavily truncated. These examples motivate compatibility-aware aggregation while avoiding a causal interpretation of attention weights.

\subsection{NuScenes Localization}

\begin{table}[t]
\centering
\caption{NuScenes 3D pedestrian-center localization. Params/Size report localizer FP32 cost; Peak/GFLOPs follow the resource protocol. FCOS3D performs its own detection; social models use $K=4$.}
\label{tab:nuscene_localization}
\scriptsize
\setlength{\tabcolsep}{2pt}
\begin{tabular}{@{}>{\raggedright\arraybackslash}m{0.29\columnwidth} >{\centering\arraybackslash}m{0.07\columnwidth} *{5}{>{\centering\arraybackslash}m{0.10\columnwidth}}@{}}
\toprule
\textbf{Localization Model} & \textbf{Social} & \textbf{Params (M)} & \textbf{Size (MB)} & \textbf{Peak (MB)} & \textbf{GFLOPs} & \textbf{ADE $\downarrow$ (m)} \\ \midrule
\multicolumn{7}{@{}>{\raggedright\arraybackslash}p{0.96\columnwidth}@{}}{\cellcolor{black!5} \textbf{Detector:} DETRPose-S (11.5 M params; 46.9 MB; 32.42 GFLOPs)} \\
\multicolumn{7}{@{}>{\raggedright\arraybackslash}p{0.96\columnwidth}@{}}{\cellcolor{black!5} Peak: 146.7 MB; coverage: 90.14\%; FCOS3D uses its own detector} \\ \midrule
MonoLoco \cite{bertoni2019monoloco} & \ding{55} & 0.41 & 1.6 & 11.3 & 0.004 & 1.450 \\
MonoTransmotion \cite{luan2025unified} & \ding{55} & 5.29 & 21.2 & 30.8 & 0.053 & 1.407 \\
UniDepth V2 ViT-L \cite{piccinelli2025unidepthv2} & \ding{55} & 353.8 & 1415.3 & 2654.1 & 2650.56 & 0.935 \\
FCOS3D R101 \cite{wang2021fcos3d} & \ding{55} & 55.0 & 219.8 & 621.3 & 682.50 & \textbf{0.901} \\ \hdashline \noalign{\vskip 0.5ex}
\textbf{MoSoLoco-MLP} & \ding{51} & 0.67 & 2.7 & 12.4 & 0.008 & 1.219 \\
\textbf{MoSoLoco-Trans} & \ding{51} & 4.62 & 18.5 & 29.8 & 0.468 & 1.130 \\ \bottomrule
\end{tabular}
\end{table}

On NuScenes (Table~\ref{tab:nuscene_localization}), social MoSoLoco-MLP/Trans reach 1.219/1.130 m ADE versus 1.450 m for MonoLoco and 1.407 m for MonoTransmotion, preserving the pose-based advantage after dataset-specific pose adaptation and localizer training. UniDepth and FCOS3D obtain lower ADE (0.935/0.901 m) over 2,253/2,372 targets (90.26/95.03\% coverage), but at much larger compute: including DETRPose-S, complete MoSoLoco-MLP/Trans contain 12.17/16.12 M parameters versus 365.3 M for UniDepth+DETRPose-S and 55.0 M for FCOS3D. Their different pipelines make this an accuracy--resource comparison rather than a controlled ablation.

\subsection{Downstream Trajectory Forecasting}

\begin{table*}[t]
\centering
\caption{Controlled JRDB trajectory forecasting. MoSoLoco uses $K=4$. OmniTraj reports best-of-20 minADE with FDE from the same mode. Each detector block uses its common retained cohort; filtering uses only observed history.}
\label{tab:traj_forecasting_gt}
\setlength{\tabcolsep}{6pt}
\begin{tabular}{@{}lcccccccc@{}}
\toprule
\multirow{3}{*}{\textbf{Localization Model}} & \multicolumn{4}{c}{\textbf{Vanilla Transformer}} & \multicolumn{4}{c}{\textbf{OmniTraj \cite{gao2025omnitraj}}} \\ \cmidrule(lr){2-5} \cmidrule(l){6-9}
 & \multicolumn{2}{c}{\textbf{Raw Forecast}} & \multicolumn{2}{c}{\textbf{Filtered}} & \multicolumn{2}{c}{\textbf{Raw Forecast}} & \multicolumn{2}{c}{\textbf{Filtered}} \\ \cmidrule(lr){2-3} \cmidrule(lr){4-5} \cmidrule(lr){6-7} \cmidrule(l){8-9}
 & \textbf{ADE $\downarrow$} & \textbf{FDE $\downarrow$} & \textbf{ADE $\downarrow$} & \textbf{FDE $\downarrow$} & \textbf{ADE $\downarrow$} & \textbf{FDE $\downarrow$} & \textbf{ADE $\downarrow$} & \textbf{FDE $\downarrow$} \\ \midrule
\multicolumn{9}{@{}l}{\cellcolor{black!5} \textbf{Detector:} DETRPose-S (Small)} \\ \midrule
GT-History Input & 0.082 & 0.144 & -- & -- & 0.059 & 0.116 & -- & -- \\ \midrule
MonoLoco \cite{bertoni2019monoloco} & 0.283 & 0.364 & 0.244 & 0.304 & 0.250 & 0.271 & 0.218 & 0.226 \\
MonoTransmotion \cite{luan2025unified} & 0.280 & 0.359 & 0.243 & 0.300 & 0.240 & 0.260 & 0.207 & 0.208 \\
\textbf{MoSoLoco-MLP ($K=4$)} & 0.259 & 0.347 & 0.215 & 0.278 & 0.226 & 0.265 & 0.182 & \textbf{0.201} \\
\textbf{MoSoLoco-Trans ($K=4$)} & \textbf{0.250} & \textbf{0.336} & \textbf{0.206} & \textbf{0.267} & \textbf{0.219} & \textbf{0.251} & \textbf{0.181} & 0.207 \\ \midrule
\multicolumn{9}{@{}l}{\cellcolor{black!5} \textbf{Detector:} DETRPose-N (Nano)} \\ \midrule
GT-History Input & 0.082 & 0.146 & -- & -- & 0.062 & 0.122 & -- & -- \\ \midrule
MonoLoco \cite{bertoni2019monoloco} & 0.310 & 0.411 & 0.256 & 0.318 & 0.264 & 0.282 & 0.221 & 0.224 \\
MonoTransmotion \cite{luan2025unified} & 0.337 & 0.448 & 0.276 & 0.346 & 0.272 & 0.281 & 0.238 & 0.243 \\
\textbf{MoSoLoco-MLP ($K=4$)} & 0.280 & \textbf{0.377} & 0.230 & 0.295 & 0.238 & 0.267 & 0.199 & 0.222 \\
\textbf{MoSoLoco-Trans ($K=4$)} & \textbf{0.275} & 0.382 & \textbf{0.216} & \textbf{0.286} & \textbf{0.223} & \textbf{0.259} & \textbf{0.187} & \textbf{0.215} \\ \bottomrule
\end{tabular}
\end{table*}

Table~\ref{tab:traj_forecasting_gt} shows lower forecasting ADE from MoSoLoco observations than from both pose-based baselines for both detectors and forecasters. For DETRPose-S raw histories, MoSoLoco-Trans gives 0.250 m Vanilla ADE and 0.219 m OmniTraj ADE versus 0.283/0.250 m for MonoLoco and 0.280/0.240 m for MonoTransmotion. With DETRPose-N, the corresponding MoSoLoco-Trans values are 0.275/0.223 m versus 0.310/0.264 and 0.337/0.272 m. Filtering reduces ADE for every predicted-history row, while the GT-history references remain substantially lower.

\begin{figure}[t!]
    \centering
    \includegraphics[width=0.9\linewidth]{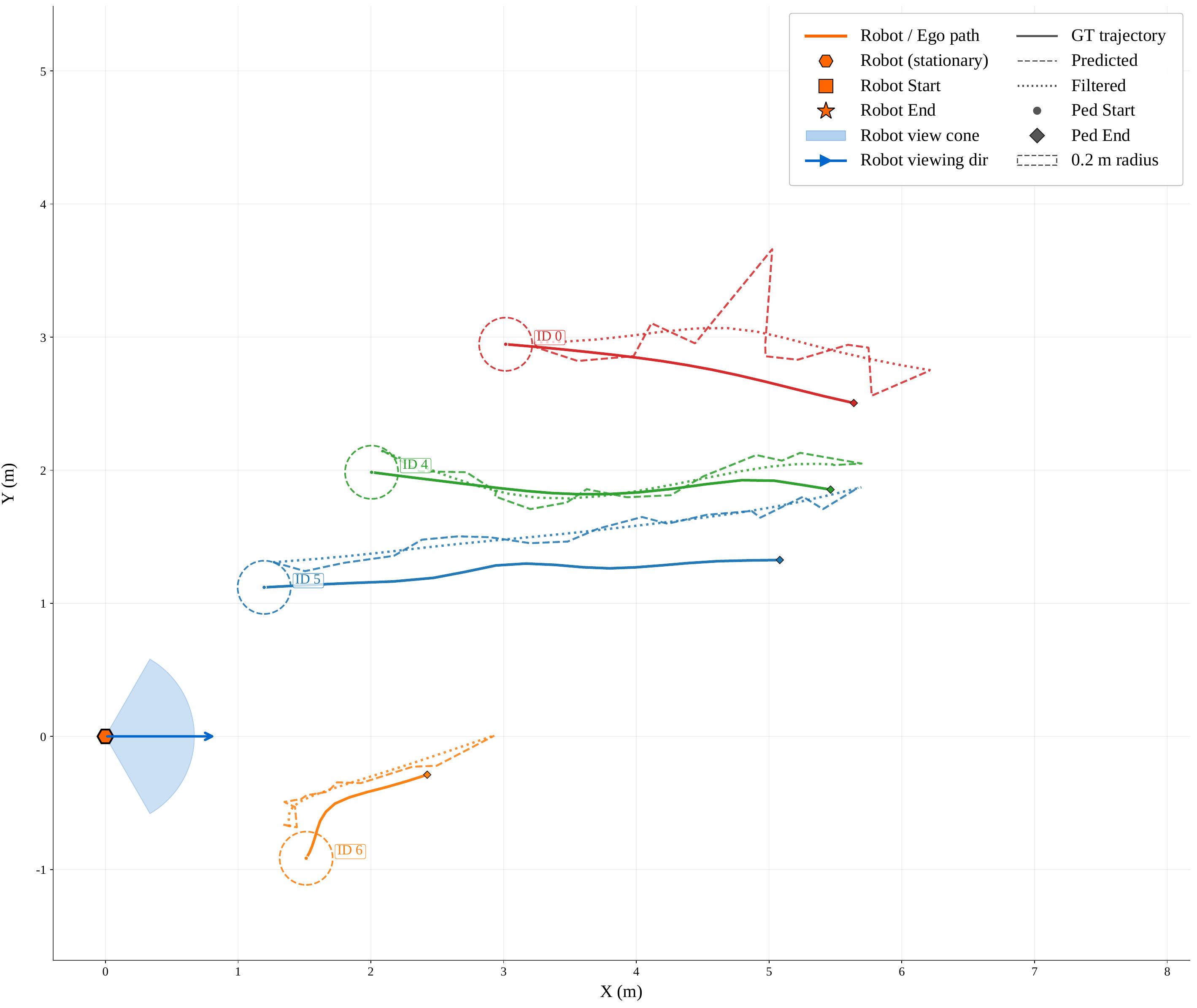}
    \caption{JRDB forecasting example. Dashed trajectories are raw localization observations; dotted trajectories apply the future-blind filter only to observed history.}
    \label{fig:birdseye}
\end{figure}

\subsection{Embedded Deployment}
\begin{table}[t]
\centering
\caption{Jetson AGX Xavier model-inference results on JRDB.}
\label{tab:jetson_deployment}
\setlength{\tabcolsep}{4pt}
\begin{tabular}{@{}llcc@{}}
\toprule
\textbf{Model} & \textbf{Runtime} & \textbf{ADE $\downarrow$} & \textbf{FPS $\uparrow$} \\ \midrule
\multirow{2}{*}{MoSoLoco-Trans} & Baseline & 0.248 & 6.31 \\
 & Optimized & 0.248 & \textbf{23.00} \\ \midrule
\multirow{2}{*}{MoSoLoco-MLP} & Baseline & 0.254 & 6.97 \\
 & Optimized & 0.254 & \textbf{24.15} \\ \bottomrule
\end{tabular}
\end{table}

TensorRT raises synchronized model-inference throughput from 6.31 to 23.00 FPS for MoSoLoco-Trans and from 6.97 to 24.15 FPS for MoSoLoco-MLP ($3.65\times$/$3.47\times$), with ADE unchanged at 0.248/0.254 m. Optimized mean latencies are 43.48/41.40 ms (95th percentile: 50.21/48.04 ms); CUDA NMS preserves query-selection masks and coverage on all 1,658 images.

\section{Discussion}

The central result is that intermediate pose-detector representations improve a lightweight pose-based monocular localizer when fused with camera-normalized skeleton geometry. The gain appears with both DETRPose-S and DETRPose-N on JRDB and remains present on NuScenes after dataset-specific pose adaptation and localizer training. The latent-only ablation is also informative: detector features can support localization by themselves with DETRPose-S, but not consistently with DETRPose-N, while the combined models outperform their latent-only counterparts for both detector capacities. Together with the feature probe, these results are consistent with the reused representations carrying localization-relevant information that complements the final 2D keypoints; they do not identify which visual cues are responsible.

Same-frame multi-person attention provides an additional but smaller refinement. Its point estimate improves for both MLP and Transformer localizers and for every tested value of $K$ relative to the no-social setting. The non-monotonic neighborhood ablation and the qualitative cases in Fig.~\ref{fig:social_examples} indicate, however, that neighboring hypotheses are not uniformly useful. The failure example is consistent with context becoming less reliable when the anchor differs strongly from surrounding upright pedestrians, whereas context can help when the anchor itself is heavily truncated. This motivates compatibility-aware aggregation using pose, visibility, or activity-related cues rather than assuming that more neighbors will always help.

The forecasting experiment evaluates downstream utility under deliberately controlled association. MoSoLoco observations yield lower forecasting ADE than the pose-based localization baselines for both a deterministic Transformer and OmniTraj, while the ground-truth-history rows remain substantially below every predicted-history condition. The result is consistent with localization quality being an important source of downstream forecasting error in the evaluated setting. It is not an end-to-end tracking result: identities are supplied through ground-truth association, failed detections remain masked, and forecasting metrics are computed on a common retained cohort. Per-frame and forecast-window coverage therefore remain necessary companion metrics when interpreting ADE and FDE.

Several limitations bound the claims. First, NuScenes is not a zero-shot transfer experiment because both the pose front end and localizers use the NuScenes training scenes before validation. Second, UniDepth and FCOS3D follow different detection and processing pipelines and obtain lower ADE than MoSoLoco on NuScenes; those rows therefore quantify an accuracy--compute trade-off rather than a controlled accuracy comparison. Third, the tables report point estimates without repeated-seed uncertainty, so small differences, including the few-centimeter social gains and the gap between the two MoSoLoco backbones, should be interpreted cautiously. Finally, Jetson measurements exclude image preprocessing, host-to-device transfer, and final post-processing. The reported 23--24 FPS is detector--localizer model-inference throughput, not full sensor-to-output latency.

\section{Conclusion}

  This work introduced \textbf{MoSoLoco}, a lightweight per-frame approach to monocular 3D human localization that reuses intermediate visual representations from a
  frozen 2D pose detector. MoSoLoco combines these representations with skeleton geometry, while a sparse same-frame interaction module uses nearby detections to
  further refine each estimate. The feature analysis showed that the detector representations retain depth-related structure beyond coarse skeleton cues, providing
  empirical support for their use in localization.

  Across JRDB and NuScenes, this design consistently improved over the evaluated keypoint-based localizers across detector capacities and fusion architectures. The
  best JRDB configuration reduced ADE by 23.9\% and achieved accuracy comparable to much larger monocular depth and 3D detection systems. More accurate locations also
  reduced error in the controlled trajectory-forecasting study, while TensorRT deployment reached 23.0--24.2 FPS on a Jetson AGX Xavier without changing localization
  accuracy. Together, these results show that representations already computed for pose estimation can improve the accuracy--efficiency balance of monocular human
  localization. Extending the approach with tracking, studying cross-domain transfer, and measuring complete sensor-to-output latency remain important directions.

\section{Acknowledgment}
The authors gratefully acknowledge the financial support provided by the Swiss National Science Foundation (SNSF) through grant no. 10003100.

\printbibliography
\end{document}